\def\year{2022}\relax
\documentclass[letterpaper]{article} 
\usepackage{aaai22}  
\usepackage{times}  
\usepackage{helvet}  
\usepackage{courier}  
\usepackage[hyphens]{url}  
\usepackage{graphicx} 
\usepackage{natbib}  
\usepackage{caption} 
\DeclareCaptionStyle{ruled}{labelfont=normalfont,labelsep=colon,strut=off} 
\usepackage{algorithmic}
\usepackage{graphicx}
\usepackage{textcomp}
\usepackage{xcolor}

\usepackage{algorithm}
\usepackage{subfigure}

\usepackage{diagbox}
\usepackage{threeparttable}
\usepackage{multirow}
\usepackage{color}
\usepackage{amsmath}
\usepackage{url}
\usepackage{booktabs}

\usepackage{setspace}

\usepackage{newfloat}
\usepackage{listings}
\floatstyle{ruled}
\newfloat{listing}{tb}{lst}{}
\floatname{listing}{Listing}
\title{Longitudinal Fairness with Censorship}

\author{
	Wenbin Zhang \normalfont{and }\bf
	Jeremy C. Weiss
}
\affiliations{
	Carnegie Mellon University\\
	wenbinzhang@cmu.edu, jeremyweiss@cmu.edu
}

\usepackage{bibentry}

\begin{document}

\maketitle

\begin{abstract}
	
	Recent works in artificial intelligence fairness attempt to mitigate discrimination by proposing constrained optimization programs that achieve parity for some fairness statistic. Most assume availability of the class label, which is impractical in many real-world applications such as precision medicine, actuarial analysis and recidivism prediction. Here we consider fairness in longitudinal right-censored environments, where the time to event might be unknown, resulting in censorship of the class label and inapplicability of existing fairness studies. We devise applicable fairness measures, propose a debiasing algorithm, and provide necessary theoretical constructs to bridge fairness with and without censorship for these important and socially-sensitive tasks. Our experiments on four censored datasets confirm the utility of our approach.
	
\end{abstract}

\section{Introduction}

With the rise of big data, artificial intelligence (AI)-based decision making systems are used in a growing number of applications, including many high-impact areas such as healthcare, employment, credit lending and criminal justice~\cite{beutel2019putting,meyer2018amazon,liu2021mitigating}. There is concern
that automated decisions made in this fashion encode and even exacerbate existing real-world disparities, inflicting harm to certain individuals or social groups~\cite{vasudevan2020lift,li2021time}. 
This issue has motivated a number of approaches to quantify and mitigate algorithmic unfairness and has given rise to an active field of research deemed AI fairness~\cite{mehrabi2021survey}. The vast majority of existing studies tackle the problem by taking the existing notions of algorithmic fairness that focus on either individual level fairness, e.g., disparate treatment, to guarantee similar people are treated similarly~\cite{dwork2012fairness}, or group level fairness, e.g., disparate impact, seeking approximate parity of some statistic across different groups~\cite{zhang2019faht}. Moreover, most of the work in this literature studies how to make machine learning algorithms fair in the presence of a class label---where the fairness notions are defined based on the class label,
and where the predictive model is trained contingent upon them. Surprisingly, little attention has been given to \emph{censorship} settings in which the time to an event of interest could be inaccessible to the learner~\cite{turner2022longitudinal}, thus making existing fairness notions and approaches inapplicable. In practice, \emph{e.g.}, evaluating the retention rates of each marketing channel in marketing analytics, and judging defendant's criminal recidivism for bail and sentencing in recidivism prediction instruments, the latter is often the case.


In this work, we consider such a censorship setting where the true time to event might be unknown to the learner, while fulfilling the requirements of fair and accurate predictions. Addressing unfairness and censorship simultaneously presents unique challenges, and transferring from respective domains is not straightforward. In contrast to previous works~\cite{verma2018fairness}, our goal here is to prevent cases of unequal treatment according to their certain characteristics or sensitive attributes (\emph{e.g.}, gender or race) at the individual and group level.
In addition, unlike previous works limited to categorical sensitive attributes that are binary~\cite{zhang2021fair}, our goal also generalizes to an k-way categorical and continuous variables. Our definitions of fairness can therefore be thought of versatile notions focusing on individual and group levels, and inclusive of data with censored individuals.


Our formulation of fairness is motivated by the following observation: in healthcare, the impressive human-level or even surpassing human-level performance of AI systems is balanced against a plethora of observed discriminatory incidents~\cite{rajkomar2018ensuring}. As an example, a Propublica report found that state-of-the-art clinical prediction models underperformed on black patients even when a treatment was aimed at a particular type of cancer that disproportionately impacted them~\cite{chen2020ethical}. Such observations also go beyond the medicinal domain with examples in marketing analytics~\cite{chang2021ethical}, actuarial~\cite{frezal2019fairness} and recidivism prediction instruments~\cite{angwin2016there}, where the common challenge, other than ethical concerns, is the individual's true time to event might be unknown. We remark that thus far the fairness-in-AI community has primarily focused on \emph{no censorship} settings with clearly defined class labels of instances, despite these prevailing censorship scenarios.


Armed with this broader observation of AI unfairness, care must be taken to ensure that an automated decision making system is fair or independent of harmful and sensitive attributes-based stereotypes in the presence of censorship. This motivates the definitions of fairness involving censored individuals and a corresponding algorithm to address discrimination with censorship that we propose in this work. More specifically, the novelty of this research comes from four aspects: i) We study a novel problem of longitudinal fairness with censorship, which is commonly rooted in socially sensitive applications but remains highly under-explored. ii) Corresponding fairness notions explicitly considering censorship are devised to measure bias in the presence of censorship, as well as a respective fair learner to ensure accurate predictions while also preserving a low discrimination score in longitudinal censorship settings. iii) We theoretically establish the connections of fairness in censored and non-censored settings, offering greater understanding and explanation for AI fairness. iv) Detailed experimental evaluations validating our model with regards to both fairness and accuracy on four real-world censored and discriminated datasets.

\section{Background and Related Work}

\subsection{Longitudinal Biased Data with Censorship}

In the typical AI fairness settings, the biased data $X$ normally consists of a sequence of feature represented instances $x_1, x_2, \cdots, x_n$. Among the feature representation, a special attribute $G$ is referred as the \emph{sensitive attribute} and its attribute values distinguish the discriminated community, i.e., the deprived group, from the privileged community, i.e., the favored group. In addition, instances are also described by their corresponding class labels $y_1, y_2, \cdots, y_n$. However, class labels can become  inaccessible in the presence of censorship. 

The discriminated and censored data, in contrast to the typical data representation, therefore further contains the survival time $T$ and an event indicator $\delta$ in addition to the observed features $x$, typically represented in the form of ($x$, $T$, $\delta$). If the event of interest has occurred, $T$ is the actual time from the individual entered the study till the time of the event occurring, and $\delta$ becomes 1 indicating certainty on the event observation; otherwise $T$ corresponds to the elapsed time between individual entered the study and last follow-up with the individual, and the event indicator $\delta$ = 0, i.e., the survival time is censored~\cite{wang2021harmonic}. 

Compared with AI fairness in supervised settings, addressing discrimination bias in censoring settings leads to censorship on $y_1, y_2, \cdots, y_n$ which limits the applicability of the existing fairness notions. In addition, the  uncertainty on $y_1, y_2, \cdots, y_n$ could also further accompany and complicate the biased decision regions. Given the discriminated and censored data $X$, the aim of longitudinal AI fairness with censorship is then to model a fair survival function $H(\cdot)$ which makes accurate predictions based on $X$ but also does not discriminate with respect to $G$ for the discriminated and censored datasets.

\subsection{AI Fairness}

While artificial intelligence is increasingly permeating facets of life, significant concerns on the unfair and discriminatory manner of AI-based systems have been voiced and observed~\cite{beutel2017data}. The AI community has responded by proposing a growing body of fairness notions to measure the level of discrimination along with a number of approaches to mitigate bias in order to provide fair decision making systems~\cite{hajian2016algorithmic,mehrabi2021survey,zhang2022fairness}.

The broad set of existing mathematical formulations of fairness can be typically divided into two main families, \emph{individual fairness} and \emph{group fairness}. The former aims to ensure that similarly situated individuals are treated similarly~\cite{dwork2012fairness} while the latter asks for group level approximate parity of some statistic over class labels~\cite{verma2018fairness}. Although a vast of fairness notions exist, most of them formulate fairness depend on class label thus limiting their applicability in censorship settings. Kamrun et al.~\cite{keya2021equitable} directly extend the existing fairness notions to the application with censoring problems, and it is the only relevant work to the best of our knowledge. However, their definitions exclude the censorship information when measuring discrimination which could introduce substantial bias as censored information can be of importance and cannot simply be ignored~\cite{clark2003survival}.

The aforementioned fairness definitions could be directly used or slightly modified as a constraint or a regularizer to enforce fairness, leading to three categories of debiasing mechanisms: \emph{pre-processing approaches}~\cite{kamiran2009classifying,vzliobaite2017measuring}, \emph{in-processing solutions}~\cite{zhang2019faht,geden2021fair}, and \emph{post-processing techniques}~\cite{hardt2016equality,fish2016confidence}. The critical limitation of these methods as well as other existing fairness works is that they are in need of class label for their unfairness formulations and algorithmic solutions, and fairness when some class labels are unknown has not been well explored~\cite{keya2021equitable}. Our work seeks to alleviate such limitation by jointly addressing bias reduction and censoring management.

\subsection{Survival Analysis}

The critical challenge of the main outcome under assessment could be unknown for a portion of the study group, deemed censorship, hinders the use of many methods of analysis. This motivates the study of \emph{survival analysis} to address the problems of partial survival information access from the study cohort~\cite{clark2003survival}. The censored data, also known as \emph{survival data}, are generally considered and modeled in terms of two quantitative terms, namely the hazard function and the survival function. The former models the instantaneous rate of event occurs at a specified time $t$ conditioned on surviving to $t$: 

\vspace{-0.1cm}
\begin{equation}
	h(t|x) = \lim\limits_{\bigtriangleup t \rightarrow 0} \frac{Pr(t<T<t+\bigtriangleup t| T\geq t, x)}{\bigtriangleup t}
\end{equation}

\noindent The latter is the probability that the event does not occur up to time $t$ and can be determined from the hazard function (and vice versa):

\begin{equation}
	S(t|x) = exp(-H(t|x)),~H(t|x)= \int_{0}^{t} h(t|x)dt
\end{equation}

Given the ubiquity of censored data in real-world applications, survival analysis has gained its popularity in various applications ranging from medicine to customer and actuarial analytics to predictive maintenance in mechanical operations~\cite{de1999mixture}. Among the various methods proposed for modeling censored data, the Cox proportional hazards model (CPH)~\cite{cox1972regression} is the most commonly used in which the multiplicative relation between the risk, as expressed by the baseline hazard function, and covariates is described. More recently, deep neural network structure has also been extended to model the feature interactions of survival data. For example, DeepSurv~\cite{katzman2018deepsurv} employs the loss function of CPH with L2 regularization to train the networks. Another line of effort is the tree based methods~\cite{bou2011review}, particularly random forests due to its superior capabilities in handling nonlinear effect of variables and avoiding restrictive assumptions such as that of proportional hazards~\cite{ishwaran2008random}. A comprehensive literature survey covering recent censored data modeling effort is provided in~\cite{wang2019machine}.

With the popularity of survival models, care must be taken to ensure their fairness, the same as other AI approaches. Our work situates in this under-explored research direction to tackle fairness in the presence of censorship. Our in-processing approach incorporates a pairwise-comparison fairness notion in the algorithm design to guide a accuracy-driven as well as fairness-oriented learning procedure. Relevantly, the survival model is modified to ensure fair risk predictions as in~\cite{keya2021equitable}. Three key differences are that our model: i) does not necessitate a distance metric to be specified, ii) explicitly considers survival information to address discrimination in the presence of censorship, and iii) enjoys the merit of free from hyperparameter tuning.

\section{Our Approach}


\subsection{Defining Bias with Censorship}

\subsubsection{Concordance Imparity}

The presence of censorship in data limits the applicability of commonly used fairness definitions introduced in the existing AI fairness studies. To fill this gap, we introduce \emph{Concordance Imparity (CI)} to specifically account for model unfairness in the presence of censorship. Specifically, CI first considers individual level pairwise comparison based on the consistency between model prediction and true outcomes, then measures, at the group level, whether the discriminative ability of the model is fairly distributed across different groups. Different from the previous definitions~\cite{keya2021equitable}, the survival time and survival information are explicitly involved in CI to avoid important information loss and introducing substantial bias. The sketch of Concordance Imparity is shown in Algorithm~\ref{alg: ci}.

\begin{algorithm}[!htb]
	\caption{Concordance Imparity}
	\label{alg: ci}
	\renewcommand{\algorithmicrequire}{\textbf{Input:}}
	\renewcommand{\algorithmicensure}{\textbf{Output:}}
	\newcommand{\continue}{\textbf{continue}}
	\begin{spacing}{0.6}
		\begin{algorithmic}[1]
			\small
			\REQUIRE Censored and biased dataset $D$, risk scores $r$,\\
			~~~~~~sensitive attribute $G$\\
			
			\ENSURE CI score
			\IF{G is continuous}
			\STATE Discretize G according to Equation~(\ref{equ:fsd})
			\ENDIF
			
			\FOR {each instance $d_i$ in $D$}
			\FOR {each instance $d_j$ in $D$ \& $d_j$ $\neq$ $d_i$}
			\IF{$t_i < t_j$ \& $\delta_i ==0~ |~ t_j < t_i$ \& $\delta_j == 0~|$ \\
				($t_i$ == $t_j$ \& ($\delta_i$ == 0 \& $\delta_j$ == 0))}
			\STATE \continue
			\ELSE
			\STATE $P_{G(d_i)==g}$ = $P_{G(d_i)==g} $+ 1
			\ENDIF
			
			\IF{$t_i < t_j$}
			\IF{$r(d_i)> r(d_j)$}
			\STATE $C_{G(d_i)==g}$ = $C_{G(d_i)==g}$+ 1
			\ELSIF{$r(d_i)= r(d_j)$}
			\STATE $C_{G(d_i)==g}$ = $C_{G(d_i)==g}$+ 0.5
			\ENDIF
			
			\ELSIF{$t_i > t_j$}
			\IF{$r(d_i)< r(d_j)$}
			\STATE $C_{G(d_i)==g}$ = $C_{G(d_i)==g}$+ 1
			\ELSIF{$r(d_i)= r(d_j)$}
			\STATE $C_{G(d_i)==g}$ = $C_{G(d_i)==g}$+ 0.5
			\ENDIF
			
			\ELSIF{$t_i == t_j$}
			\IF{$\delta_i == 1~\&~\delta_j== 1$}
			\IF{$r(d_i)== r(d_j)$}
			\STATE $C_{G(d_i)==g}$ = $C_{G(d_i)==g}$+ 1
			\ELSE	
			\STATE $C_{G(d_i)==g}$ = $C_{G(d_i)==g}$+ 0.5
			\ENDIF
			
			\ELSIF{$\delta_i == 0~\&~\delta_j == 1 ~\&~ r(d_i)< r(d_j)$}
			\STATE $C_{G(d_i)==g}$ = $C_{G(d_i)==g}$+ 1
			\ELSIF{$\delta_i == 1~\&~\delta_j == 0 ~\&~ r(d_i)> r(d_j)$}
			\STATE $C_{G(d_i)==g}$ = $C_{G(d_i)==g}$+ 1
			\ELSE 
			\STATE $C_{G(d_i)==g}$ = $C_{G(d_i)==g}$+ 0.5
			
			\ENDIF
			
			\ENDIF
			\ENDFOR
			\ENDFOR
			
			\STATE CF(G=g) = $C_g$/$P_g$
			\RETURN CI= $\max_{g, g' \in G~\&~g\neq g'}|CF_{(g)}- CF_{(g')}|$
		\end{algorithmic}
	\end{spacing}
	
\end{algorithm}

The concordance imparity measurement starts with deciding whether the sensitive attribute $G$ is an continuous attribute (line 1) and discretizes G according to the proposed \textit{fair survival difference} to be discussed in the following section if so (line 2). Next, CI forms all possible pairs of comparison for each individual and omits those incomparable pairs, \emph{i.e.}, the shorter time is censored, and both-censored pairs with identical survival time (line 4-7). The remaining are the permissible pairs across different demographic groups (line 8-9). Among them CI checks three possibilities: 1) if the individual under consideration, $d_i$, has a shorter survival time, $t_i$, than the compared individual's survival time $t_j$, then the concordance count of respective demographic group, $C_{G(d_i)==g}$, that $d_i$ belongs to increments by 1 if the model actually assigns a higher risk score to $d_i$ and by 0.5 if predicted outcomes are tied (line 11-16). 2) Line 17-22 checks the opposite scenario that $d_i$'s survival time is longer than $d_j$'s and counts are incremented similarly. 3) When identical survival times observed (line 23) and neither are censored (line 24), $d_i$'s respective concordance count will be added by 1 on the condition that the predicted outcomes are tied, and by 0.5 otherwise (line 25-29); When the survival times are still the same but not both are censored, $C_{G(d_i)==g}$ increments by 1 if the non-censored individual has a higher predicted risk score and by 0.5 otherwise (line 30-36). Line 40 then evaluates \emph{concordance fraction (CF)}, the group-wise correct pairwise ordering, and the final CI score is measured as the largest deviation of discriminative abilities across different demographic groups of the model (line 41). The lower the concordance imparity score the fairer the model.

Note that in comparison to existing fairness notions mainly focus on binary protected categorical attributes~\cite{verma2018fairness}, other than the explicit inclusion of censorship information, concordance imparity also looks at the generalization of measuring the level of discrimination to k-way categorical attributes. This is done by reformulating the general discrimination measurement to consider the largest difference among sub-community, which is equivalent to the typical fairness definition when $k=2$. In addition, CI also considers the discrimination in regards to continuous attribute domain. Specifically, the allowing test~\cite{han2011data} is first used to explore potential binary split candidates, then the allowed split with the largest merit achieved is selected as the threshold for splitting. The merit is gauged according to the proposed \textit{fair survival difference} to be discussed hereafter. The calculation of CI can then proceed the way as the categorical attribute after such a discretization. This gives us a generalized definition extending CI to fair regression tasks, enabling an inclusive discrimination evaluation consisting of censored individuals.

\subsubsection{Fair Calibration}

To further exploit the semantic information of survival probabilities produced by the model which are also labels for individuals~\cite{haider2020effective}, we propose \emph{fair calibration (FC)} to measure whether the model under consideration creates probability value based disparity systematically.

In summary, fair calibration of survival probabilities was assessed by: i) plotting group-wise (i.e., $\forall g \in G$) observed proportions versus predicted probabilities and ii) by calculating corresponding fair calibration validity. The first step examines the agreement between the predicted probabilities of the model with the observed outcome. To do so, FC sorts and splits the predicted probabilities for a particular time $t$ for each demographic group into deciles, then checks whether these probabilities are sufficiently close to the observed proportions. Note that smaller subgroups will lead to greater statistical uncertainty about within-group predictive and fairness performance, but provide more similarity of instances within subgroups. Fewer subgroups result in the opposite. Convention is to use deciles, which is what we follow. In addition, the observed proportions could be unknown due to the censorship. To this end, the Kaplan-Meier (KM) curve estimate~\cite{ranstam2017kaplan} is employed and the significance of these group-wise results respect to these bins are defined by the Hosmer-Lemeshow (HL) goodness-of-fit test statistic~\cite{hosmer1980goodness}: 

\begin{equation}
	HL_g(S(t|x))= \sum_{i=1}^{B} \frac{({KM_i}_g- \bar{p_i}_g)^2{n_i}_g}{\bar{p_i}_g(1-\bar{p_i}_g)}
\end{equation}

\noindent where $B$ represents the number of bins, ${KM_i}_g$ is the KM estimated probability in the $i$th decile at time $t$, $\bar{p_i}_g$ is the predicted probability for individuals in the $i$th decile and ${n_i}_g$ is the number of observations in decile $i$. Note that all of them are group-wise, i.e., $G= g$.  

Based on the group-wise examined level of agreements, the second step of FC evaluates consistency across different demographic groups as shown in Equation~(\ref{equ:fc}), involving the first two fair calibrated scenarios and the third biased calibrated scenario otherwise: i) representation consistency: the predicted probabilities are representative of the actual probabilities; the p-value of each group's HL statistic passes the test with a value not smaller than 0.05, ii) difference consistency: representation inconsistent but the difference between predicted probabilities and actual probabilities (i.e., $\triangle p_g/\triangle p_{g'}$) is accordant across subgroups; the p-value of each difference test among subgroups evaluated by Wilcoxon signed-rank test~\cite{woolson2007wilcoxon} is greater than 0.5, iii) Neither representation nor difference is consistent.

\begin{equation}
	\label{equ:fc}
	FC= 
	\begin{cases}
		\text{fair~calibrated,} & p(HL_g(S(t|x)))\geq 0.05\\
		& ~~~~~~~~~~~~~~~~~~~~~~~~~~~~~~~~~~~\forall g\in G\\
		\text{fair~calibrated,} & wilcoxon(\triangle p_g, \triangle p_{g'})> 0.5\\
		& ~~~~~~~~~~~~~~~~~~~~~~~~~~~~~~~\forall g, g' \in G\\
		\text{biased~calibrated,} & \text{otherwise}
	\end{cases}
\end{equation}

\subsection{Mitigating Bias with Censorship}
\label{sec:fsrf}


Armed with the afore established fairness statistics, measuring unfairness in the presence of censorship becomes feasible. This section then serves to fulfill the subsequent bias mitigation amidst censorship. 


The proposed approach follows the general idea of random forests (RF) by constructing an array of base learners to improve the predictive ability. In particular to censored data, RF is also nonparametric while enjoying the merits of nonlinear interactions modeling~\cite{wang2019machine}. However, such ensemble methods aim to optimize for data encoding for predictive performance, and fairness, which we desired to add, is imperceptible~\cite{ishwaran2008random}. In this work, to jointly optimizing for censored data encoding and debiasing, we propose \emph{Fair Survival Random Forests (FSRF)} which extends the RF model in two ways: i) by introducing a new splitting criterion that jointly considers the reduction of an attribute split w.r.t. impurity and also w.r.t. discrimination, ii) by illustrating the way to provide fair risk predictions amidst censorship.


The information gain and Gini impurity, when censorship is absent, are commonly used splitting criteria to guide the induction of the tree for classification performance~\cite{han2011data}. However, the presence of censorship leads to inaccessibility of class label thus making their computation impractical. The survival difference can be instead used to measure the impurity reduction for candidate splitting evaluation. In FSRF, such survival difference between different groups are evaluated by the \emph{logrank test}~\cite{bland2004logrank}:   



\begin{equation}
	SD= \frac{\sum_{j=1}^{k} (O_j-E_j)}{\sqrt{\sum_{j=1}^{k} V_j}} \sim N(0,1)
\end{equation} 

\noindent where $O_j$ and $E_j$ represent the observed number of events and the expected number of events, respectively with $V_j$ being the variance of $O_j$. The candidate with a larger logrank test therefore leads to more similarity within child nodes but also more dissimilarity among child nodes if it is being selected for splitting.

We then combine concordance imparity and survival difference as a conjunctive criterion that takes both predictive performance and fairness into consideration. We define the conjunctive criterion \emph{fair survival difference (FSD)} as:     

\vspace{-0.2cm}
\begin{equation}
	\label{equ:fsd}
	FSD= 
	\begin{cases}
		\log SD - \log CI & \text{if CI $\neq$ 0}\\
		+\infty & \text{otherwise}
	\end{cases}
\end{equation} 

Intuitively, FSD closely ties SD and CI. When the candidate attributes that are free of discrimination, i.e., CI equals 0, FSD becomes positive infinite to prioritize fair splitting. 

In practice, the calculation of CI depends on the distribution that a potential splitting could lead to and the associated risk predictions based on the distribution. FSRF employs the cumulative hazard function $H(t|x)$ to predict such risk score for the sake of having a direct interpretation of the expected number of events, as well as serving the intermediate function between hazard and survival functions for direct derivation when needed. Formally, the risk score is estimated by the Nelson-Aalen estimator~\cite{borgan2014n} as: 

\vspace{-0.3cm}
\begin{equation}
	H(t|x)= \sum_{j\leq t} \frac{d_j}{n_j}
\end{equation} 

\noindent where $d_j$ and $n_j$ represent the number of individuals experiencing events and have not experienced the event at time $j$ respectively, and $t$ is evaluated as the last event time. Similar to the same node sharing identical class label in non-censoring trees, all individuals within the node of FSRF share an identical risk score when evaluating splitting but also predicting final risk.

\subsection{Bridging Fairness in the Presence and Absence of Censorship}

The previously proposed fairness definitions and debiasing algorithm explicitly consider the indispensable survival information and lay the groundwork for AI fairness in the presence of censorship. In addition, it is also desire to understand the connections of AI fairness in the presence and absence of censorship to build fundamental theoretical frameworks for AI fairness. So are of practitioners and policy makers' interests to help them have an additional layer of understanding of AI fairness for fair decision making navigation and customization. This section serves to fill this gap. 

As previously discussed, the standard AI fairness techniques, such as the most widely used statistical parity, are not suitable in the presence of censoring in the data~\cite{verma2018fairness}. Here, we take the devised concordance imparity as the illustrative example to connect AI fairness in the presence and absence of censorship, so as to facilitate the study of fairness with censorship. Recall that CI first measures subgroup-wise fraction of concordant pairs, then gauges concordance difference between different subgroups defined by the sensitive attributes. The first part of CI can therefore be thought of as an extension of the standard concordance index (C-index)~\cite{li2016multi} in subgroup-wise and then along with the second part as the weighted subgroup based area under the receiver operating curve (AUC) difference in no censoring settings. We will next elaborate this connection.

\begin{table}[!htbp]
	\footnotesize
	\setlength\tabcolsep{5pt}
	\begin{tabular}{ccccc}
		\toprule
		Charac.& SUPPORT & ROSSI & COMPAS & KKBOX \\
		\midrule
		Sample \# 			   &   	8,873	& 432     &  10,325   & 2,814,735   \\
		Censored\%	& 	0.320	& 0.736   & 0.732     & 0.347   \\
		Feature \# 				& 	14		   & 9         & 14           & 18 \\
		\begin{tabular}[c]{@{}c@{}}Sensitive \\ Attribute	\end{tabular} 	&  	gender	& race      & race    & gender \\
		\begin{tabular}[c]{@{}c@{}}Sensitive  \\ Value \end{tabular} 		 &   female	 & \begin{tabular}[c]{@{}c@{}}African\\ American \end{tabular}    &   \begin{tabular}[c]{@{}c@{}}African\\ American \end{tabular}    & female \\ \bottomrule
	\end{tabular} 
	\caption{An overview of the datasets.} 
		\label{tab:dataset_info}
\end{table}

\vspace{-0.1cm}
Starting from the standard C-index, it is a ``global'' index for validating the predictive ability of the model. Specifically, it is the fraction of pairs within the whole group, where the observation experienced the event of interest had a higher risk score than an observation who experienced the event later or had not experienced the event, representing the global assessment of the model's discrimination power. By definition, the C-index is a generalization of the Wilcoxon-Mann-Whitney statistics~\cite{austin2012interpreting} and thus of the AUC with equivalence in binary classification in the absence of censorship. The difference between the concordance part of CI and C-index is that the concordance of CI measures subgroup-wise fraction of concordant pairs in comparison to the whole group-wise concordant probability of C-index. Note that the subgroup-wise concordance calculation of CI compares an observation with both intra group and inter group observations, i.e., with all remaining observations other than itself. The concordance of CI can therefore be regarded as a subgroup based AUC, abbreviated as ${s}_{}$AUC, in contract to the standard global AUC of C-index. Next, based on the obtained ${s}_{}$AUC from different subgroups, the imparity part of CI gauges concordance difference among them. Finally, CI can be interpreted as the largest deviation among weighted ${s}_{}$AUC, and CI's counterpart of the original formulation (e.g., line 41 in Algorithm~\ref{alg: ci}) in the absence of censorship is:  

\vspace{-0.3cm}
\begin{equation}
	CI= \max_{g, g' \in G~\&~g\neq g'}|{s}_{}AUC_{(g)}- {s}_{}AUC_{(g')}|
\end{equation}  

\noindent where ${s}_{}AUC_{(G=g)}$ is the weighted subgroup based AUC from subgroup that defined by $g$ and is mathematically represented as: 

\vspace{-0.3cm}
\begin{equation}
	{s}_{}AUC_{(g)}= \frac{n_g{s}_{}AUC_g + \sum_ {j=1}^{|\{G-g\}|} n_{g'}{s}_{}AUC_{g'}}{\sum_{j=1}^{|G|} n_j} 
\end{equation}    

\noindent where ${s}_{}AUC_g$ and ${s}_{}AUC_{g'}$ are AUC values when comparing with intra- and inter-subgroup observations, respectively, and $n_j$ represents respective number of comparable pairs of each subgroup.

\begin{table*}[!htbp]
	\small
	\centering
	\begin{tabular}{ccccccc}
		\toprule
		Datasets & \diagbox{Method}{Metrics} & CI\% & Fair Calibration & C-index\% & Brier Score\% & Time-dependent AUC\%\\
		\midrule
		\multirow{6}{*}{SUPPORT} 
		& IDCPH  & 19.12  & Not fair calibrated  		  &  69.08   &  31.16    & 76.17\\
		& GDCPH &  13.12  & \textbf{Fair calibrated}    &  75.12   &  24.46	  & 78.21\\
		& CPH     &  17.45 & Not fair calibrated 		   &   74.11  &   21.21    & 80.02\\
		& RSF      &  20.11 & Not fair calibrated 			&   75.18  &  16.64		& 81.01\\
		& DeepSurv  & 18.65 & Not fair calibrated 		&   75.65  &  16.11		& 80.68\\
		& FSRF   & \textbf{9.21}  &  \textbf{Fair calibrated}   & \textbf{76.17} & \textbf{13.23}& \textbf{82.86}\\
		\midrule
		\multirow{6}{*}{ROSSI} 
		& IDCPH  &  15.31 & Not fair calibrated  		  & 52.28  & 18.73 &  77.32 \\
		& GDCPH &  9.32  & \textbf{Fair calibrated}    & 59.34 & 22.87 & 78.51 \\
		& CPH     & 11.43  & Not fair calibrated 			& 64.24 & 17.67 & 77.12\\
		& RSF      & 16.53 & Not fair calibrated 			& 65.56 & 15.12 & 79.32\\
		& DeepSurv  & 12.32 & Not fair calibrated 		& 66.67  & 14.71 & 77.17\\
		& FSRF   & \textbf{8.92}  &  \textbf{Fair calibrated}   & \textbf{69.02}  & \textbf{12.69}& \textbf{79.65}\\
		\midrule
		\multirow{6}{*}{COMPAS}  
		& IDCPH      & 25.18 & 	Not fair calibrated         & 62.16  & 25.03 & 63.78 \\
		& GDCPH   	& 11.77  &	\textbf{Fair calibrated}  & 72.16 & 16.32 &  66.21  \\
		& CPH 			& 22.43 &  Not fair calibrated	     & 69.24 & 20.35 &  65.15 \\
		& RSF 		 	& 25.32 &   Not fair calibrated       & 72.61   & 15.62 &  71.76\\
		& DeepSurv  & 16.72 &  Not fair calibrated         & 75.12 & \textbf{13.42}& 71.83\\
		& FSRF		&  \textbf{9.63}   &  \textbf{Fair calibrated} & \textbf{76.24}   & 13.81 & \textbf{72.33}\\
		\midrule
		\multirow{6}{*}{KKBOX} 
		& IDCPH   &  17.79 & Not fair calibrated		   & 72.61 				& 21.23 &  69.73 \\
		& GDCPH   & 14.98 & \textbf{Fair calibrated}   & 79.45  		   & 19.92 &  73.03 \\
		& CPH      & 18.91 &  Not fair calibrated    		 &  80.02 			 & 18.17 & 72.95 \\
		& RSF      & 21.14 &  Not fair calibrated  			  &  82.32 			   & 14.24 &  78.18\\
		& DeepSurv  & 20.66  &  Not fair calibrated 	& \textbf{83.01} & 14.33  & 80.71\\
		& FSRF   & \textbf{14.42} & \textbf{Fair calibrated} & 82.43 & \textbf{13.13}& \textbf{82.16} \\
		\midrule
	\end{tabular}
	\caption{Performance comparison of all methods on various datasets. The best results are marked in bold.} 
	\label{tab:effectiveness}
\end{table*}

\section{Experimental Results}

\subsection{Dataset Description}

We validate our model on four real-world censored datasets with socially sensitive concerns: i) The \emph{SUPPORT} dataset is from a large study to understand prognoses preferences outcomes and risks of treatment by analyzing the survival time of inpatients~\cite{knaus1995support}. ii) The \emph{ROSSI} dataset pertains to predict the reoffending risk score of convicted criminals from Maryland state prisons, who were followed up for one year after release~\cite{fox2012rcmdrplugin}. iii) The landmark algorithmic unfairness \emph{COMPAS} dataset to predict recidivism from Broward County~\cite{angwin2016there}. iv) The \emph{KKBOX} dataset from the WSDM-KKBox's Churn Prediction Challenge~\cite{kvamme2019time} to study users' risk scores of canceling their subscription from KKBOX. Table~\ref{tab:dataset_info} is a summary description of them. Note that survival time and censoring information are explicitly included in our study to specifically account for censorship.

\subsection{Comparison Methods}
We compare FSRF against five baselines to evaluate its theoretical design: i) two recent proposed fair survival models \emph{IDCPH} and \emph{GDCPH}~\cite{keya2021equitable}, which are the most competitive approaches among several variants proposed therein, and are the only works for fair survival analysis problem to the best of our knowledge, ii) along with the baseline therein, the most commonly used survival analysis tool \emph{CPH}~\cite{cox1972regression}, iii) the state of the art random forests based non-linear survival model \emph{RSF}~\cite{ishwaran2008random}, and iv) the most recent deep model on survival analysis \emph{DeepSurv}~\cite{katzman2018deepsurv}. We do not compare with other fairness baselines due to their inapplicability in the presence of censorship.  

\begin{table*}[!htbp]
	\small
	\centering
	\begin{tabular}{ccccccccccccc}
		\toprule
		\multirow{2}{*}{Datasets}& \multicolumn{2}{c}{\begin{tabular}[c]{@{}c@{}}C-index\%\\ discriminated\end{tabular}} & \multicolumn{2}{c}{\begin{tabular}[c]{@{}c@{}}C-index\%\\ privileged \end{tabular}}& \multicolumn{2}{c}{\begin{tabular}[c]{@{}c@{}}Brier Score\%\\ discriminated\end{tabular}} & \multicolumn{2}{c}{\begin{tabular}[c]{@{}c@{}}Brier Score\%\\ privileged \end{tabular}}& \multicolumn{2}{c}{\begin{tabular}[c]{@{}c@{}}Time-dependent\%\\ AUC discriminated\end{tabular}} & \multicolumn{2}{c}{\begin{tabular}[c]{@{}c@{}}Time-dependent \%\\ AUC privileged \end{tabular}}\\ 
		\cline{2-13}
		&  FSRF-  & FSRF & FSRF- & FSRF &  FSRF- & FSRF &  FSRF- & FSRF &  FSRF- & FSRF &  FSRF- & FSRF\\
		\midrule
		SUPPORT   &  60.05   &  69.64   &  80.16 & 78.85  & 25.54  &  17.65  & 8.87 & 10.14  & 73.45 & 79.67 & 87.52 &
		85.92\\
		ROSSI   	 & 57.71    & 63.78     & 74.24  & 72.7  & 21.03  &  16.36  & 9.87 & 10.22  & 69.98 & 73.77 & 84.87 & 82.21 \\
		COMPAS  & 54.82    &  68.88   &  80.14  & 78.51 &  18.76  & 16.71 &  7.66 & 11.67 & 62.81 & 65.65  & 77.62 & 75.31\\
		KKBOX     &  64.53    & 70.22   &  85.67  & 84.64 &  18.87  & 14.97 & 7.16  & 9.12 & 72.31 & 78.03 & 85.87 & 
		84.52 \\
		\bottomrule
	\end{tabular}\vspace{-0.2cm}
	\caption{\vspace{-0.3cm}Prediction performance confusion matrix for FSRF.} 
		\label{tab:confusionmatrix}
\end{table*}

\subsection{Performance Comparison}

Due to the presence of censoring in the data, the standard evaluation metrics of AI fairness such as accuracy and statistical parity are not suitable for measuring the performance in AI fairness with censorship~\cite{verma2018fairness}. Instead, in addition to the previously tailored fairness measures considering censorship, the typical survival accuracy metrics, including the C-index, Brier score and Time-dependent AUC, are utilized to evaluate the predictive performance of our model and other baselines. The \emph{C-index}~\cite{harrell1982evaluating} evaluates a model's discrimination power in terms of correct pairwise ordering, and is a generalization of the area under ROC curve (AUC) in the presence of censorship. The \emph{Brier score}~\cite{brier1951verification} is roughly the mean squared difference of the probability estimations assigned to possible outcomes and the actual outcome. Different from C-index, the lower the Brier score the merrier. Finally, the \emph{Time-dependent AUC}~\cite{chambless2006estimation} tests how well a model can distinguish individuals who experienced the event of interest from those have not prior to or at time $t$, and thus the model with a higher Time-dependent AUC score is desired. Table~\ref{tab:effectiveness} provides detailed 5-fold cross validation results of all methods on various longitudinal biased data with censorship. 

\begin{figure}[!htbp]
	\centering
		\subfigure[SUPPORT]{
			\includegraphics[height=0.2\textheight,width=0.2\textwidth]{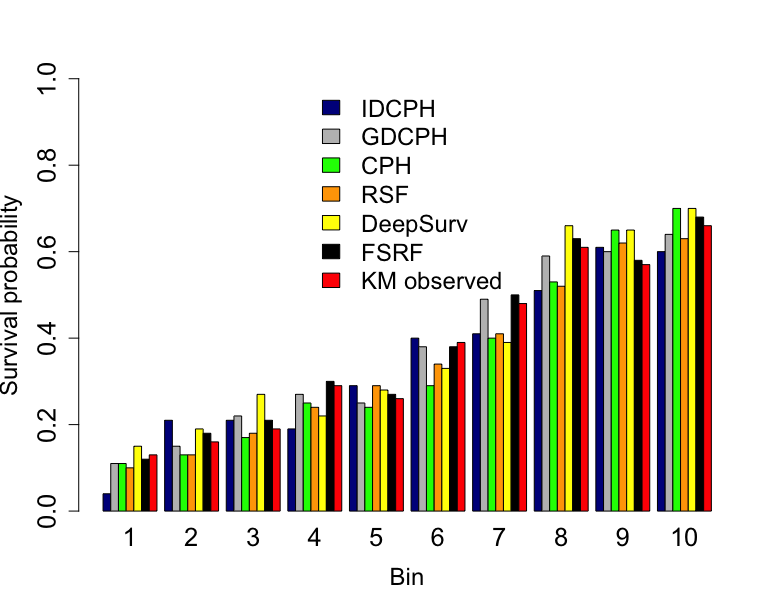}
		}\vspace{-0.4cm}
		\subfigure[ROSSI]{
			\includegraphics[height=0.2\textheight,width=0.22\textwidth]{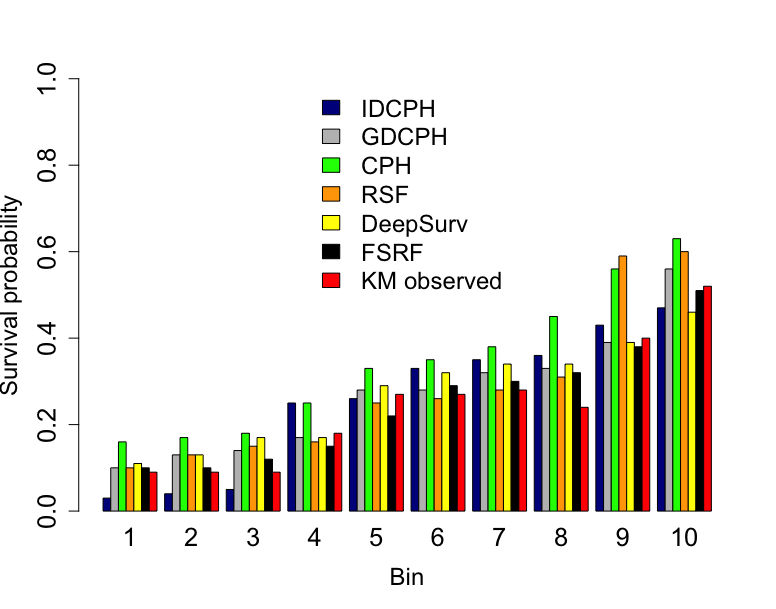}
		}
		\subfigure[COMPAS]{
			\includegraphics[height=0.2\textheight,width=0.22\textwidth]{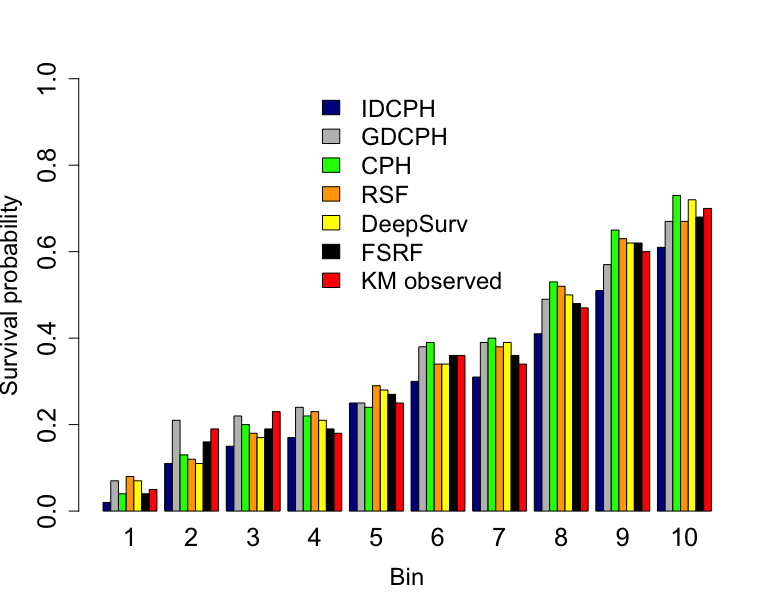}
		}%
		\subfigure[KKBOX]{
			\includegraphics[height=0.2\textheight,width=0.22\textwidth]{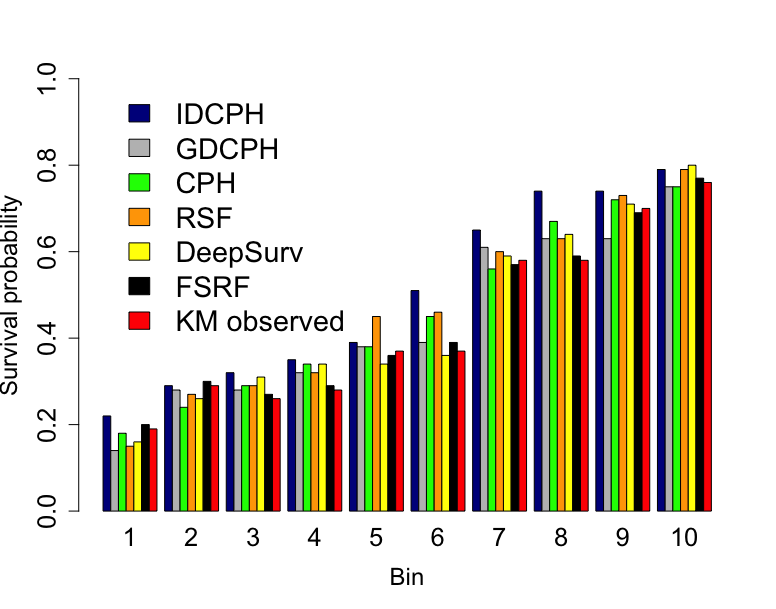}
		}%
	\vspace{-0.3cm}
	\caption{The fair calibration plots of the discriminated community. The color and height of the bar represent different methods and corresponding probabilities. The true probability observed by KM is marked in red while FSRF's is in dark.\vspace{-0cm}}
	\label{fig:fairCalibration}
\end{figure}

The results in Table~\ref{tab:effectiveness} show that the proposed model wins almost every metric across all of the datasets. This demonstrates that our proposed method has a superior debiasing capability while maintaining competence in predictive performance in the presence of censorship. Specifically, our new FSRF dominates all other methods when diminishing discrimination with censorship. This result is especially important when we contrast to other fair models proposed for censoring settings, which verifies the necessity of including survival information and survival time to mitigate bias with censorship as well as the drawback of involving task-specific similarity metric. In addition, our model, on all datasets tested, does not suffer from performance instability as other methods do, indicating FSRF is a more robust approach to building fair model with censorship. In terms of prediction performance, FSRF outperforms other models on almost all metrics in all of the datasets. This reflects that FSRF is able to handle and utilize both censored and uncensored instances when building fair model with censorship. What's more, FSRF, in contrast to other fair survival baselines, performs no parameter tuning, thus benefiting end user with simplicity while making fair decisions in the presence of censorship.

From Table~\ref{tab:effectiveness} we also note that our fairness regularizer is able to actually improve predictive performance. To have a better understanding of this phenomena, we further analyze the predictive performance confusion matrix of FSRF, according to the sensitive attribute that defines the discriminated community and privileged community as well as with and without our fairness constraints, represented by FSRF and FSRF-. Table~\ref{tab:confusionmatrix} summarizes the results. As one can see, improvement on the characterization of discriminated community is indeed achieved by including fairness attention in our method. What's more, the improved overall prediction performance also demonstrates the potential additional rewards of the debiasing design of FSRF.

Table~\ref{tab:effectiveness} additionally shows the risk prediction of FSRF is fairly calibrated as each demographic group's p-value passes its significance test, suggesting FSRF's predicted probabilities are representative of corresponding community's true probabilities. To see the full picture behind the unrankable p-values, Figure~\ref{fig:fairCalibration} graphs the predicted probabilities by FSRF in comparison to the true probabilities observed by KM (fair calibration plots of privileged community are omitted due to space constraints). In the visualization, the heights of dark bars are always close to the red bars' while bars in other colors do not follow this pattern which conclusively match with the results of fair calibration, suggesting FSRF is indeed an effective fair risk predictor.

\section{Conclusion}

Despite the increasing attention on AI fairness, existing studies have mainly focused on no censorship settings. This paper tackles fairness with censorship which is particularly prevalent in many real-world socially sensitive applications. To accomplish this objective, we devised generalized censored-specific fairness notions to quantify unfairness along with a unified debiasing algorithm to mitigate discrimination in the presence of censorship. The results on real biased datasets with censorship show our propose techniques are versatile in censoring settings. This work studies a new research problem and opens possibilities for future work on AI fairness with a broader applicability to practical scenarios concerning fairness.


\bibliography{aaai22}

\bigskip

\end{document}